\documentclass[11pt]{article}

\usepackage{xcolor}
\usepackage{graphicx}
\usepackage{xspace}
\usepackage{listings}

\lstdefinelanguage{pseudocode}{
    basicstyle=\small,
	keywordstyle=\bf \small,
	mathescape=true,
	tabsize=20,
	xleftmargin=4.0ex,
	basicstyle=\normalsize,
	sensitive=false,
	columns=fullflexible,
	keepspaces=false,
	basewidth=0.05em,
	moredelim=[il][\rm]{//},
	moredelim=[is][\sf \figuresize]{!}{!},
	moredelim=[is][\bf \figuresize]{*}{*},
	keywords={automaton, algorithm, and, 
		break,
		choose,const,continue, components,
		discrete, do,
		eff, external,else, elseif, evolve, end, each, exit,
		fi,for, forward, from, find, 
		hidden,
		in,input,internal,if,invariant, initially, imports,
		let,
		mode,
		or, output, operators, od, of,
		pre,
		return,
		such,satisfies, stop, signature, simulation, sample,
		trajectories,trajdef, transitions, that,then, type, types, to, tasks,
		variables, vocabulary, 
		when,where, with,while},
	emph={set, seq, tuple, map, array, enumeration},   
	literate=
	{(}{{$($}}1
	{)}{{$)$}}1
	{\\in}{{$\in\ $}}1
	{\\preceq}{{$\preceq\ $}}1
	{\\subset}{{$\subset\ $}}1
	{\\subseteq}{{$\subseteq\ $}}1
	{\\supset}{{$\supset\ $}}1
	{\\supseteq}{{$\supseteq\ $}}1
	{\\forall}{{$\forall$}}1
	{\\le}{{$\le\ $}}1
	{\\ge}{{$\ge\ $}}1
	{\\gets}{{$\gets\ $}}1
	{\\cup}{{$\cup\ $}}1
	{\\cap}{{$\cap\ $}}1
	{\\langle}{{$\langle$}}1
	{\\rangle}{{$\rangle$}}1
	{\\exists}{{$\exists\ $}}1
	{\\bot}{{$\bot$}}1
	{\\rip}{{$\rip$}}1
	{\\emptyset}{{$\emptyset$}}1
	{\\notin}{{$\notin\ $}}1
	{\\not\\exists}{{$\not\exists\ $}}1
	{\\ne}{{$\ne\ $}}1
	{\\to}{{$\to\ $}}1
	{\\implies}{{$\implies\ $}}1
	{<}{{$<\ $}}1
	{>}{{$>\ $}}1
	{=}{{$=\ $}}1
	{~}{{$\neg\ $}}1
	{|}{{$\mid$}}1
	{'}{{$^\prime$}}1
	{\\A}{{$\forall\ $}}1
	{\\E}{{$\exists\ $}}1
	{\\/}{{$\vee\,$}}1
	{\\vee}{{$\vee\,$}}1
	{/\\}{{$\wedge\,$}}1
	{\\wedge}{{$\wedge\,$}}1
	{=>}{{$\Rightarrow\ $}}1
	{->}{{$\rightarrow\ $}}1
	{<=}{{$\Leftarrow\ $}}1
	{<-}{{$\leftarrow\ $}}1
	{~=}{{$\neq\ $}}1
	{\\U}{{$\cup\ $}}1
	{\\I}{{$\cap\ $}}1
	{|-}{{$\vdash\ $}}1
	{-|}{{$\dashv\ $}}1
	{<<}{{$\ll\ $}}2
	{>>}{{$\gg\ $}}2
	{||}{{$\|$}}1
	{[}{{$[$}}1
	{]}{{$\,]$}}1
	{[[}{{$\langle$}}1
	{]]]}{{$]\rangle$}}1
	{]]}{{$\rangle$}}1
	{<=>}{{$\Leftrightarrow\ $}}2
	{<->}{{$\leftrightarrow\ $}}2
	{(+)}{{$\oplus\ $}}1
	{(-)}{{$\ominus\ $}}1
	{_i}{{$_{i}$}}1
	{_j}{{$_{j}$}}1
	{_{i,j}}{{$_{i,j}$}}3
	{_{j,i}}{{$_{j,i}$}}3
	{_0}{{$_0$}}1
	{_1}{{$_1$}}1
	{_2}{{$_2$}}1
	{_n}{{$_n$}}1
	{_p}{{$_p$}}1
	{_k}{{$_n$}}1
	{-}{{$\ms{-}$}}1
	{@}{{}}0
	{\\delta}{{$\delta$}}1
	{\\R}{{$\R$}}1
	{\\Rplus}{{$\Rplus$}}1
	{\\N}{{$\N$}}1
	{\\times}{{$\times\ $}}1
	{\\tau}{{$\tau$}}1
	{\\alpha}{{$\alpha$}}1
	{\\beta}{{$\beta$}}1
	{\\gamma}{{$\gamma$}}1
	{\\ell}{{$\ell\ $}}1
	{\\TT}{{\hspace{1.5em}}}3        
}

\lstdefinelanguage{pseudocodeNums}[]{pseudocode}
{
	numbers=left,
	numberstyle=\tiny,
	stepnumber=2,
	numbersep=4pt
}

\newcommand{\A}{\mathcal{A}}

\usepackage{amsmath}
\usepackage{amsfonts}
\usepackage{todonotes}

\usepackage{fullpage}
\usepackage{hyperref}
\usepackage{adjustbox}
\hypersetup{
  colorlinks=true,
  citecolor={blue},
  linkcolor = {blue},
  pagecolor={blue},
  bookmarksopen=false,
  bookmarksnumbered=true
}
\pdfoutput=1

\begin{document}
%
\title{DeepMask: an algorithm for cloud and cloud shadow detection in optical satellite remote sensing images using deep residual network}

\author{\small Ke Xu,
        Kaiyu Guan,
        Jian Peng,
        Yunan Luo,
        Sibo Wang \\ 
	\small University of Illinois at Urbana-Champaign \\
{\small \texttt {{\{kexu6, kaiyug, jianpeng, yunan, sibow2\}@illinois.edu}}}
}

%


\maketitle


%

\begin{abstract}
\label{sec:abstract}
Detecting and masking cloud and cloud shadow from satellite remote sensing images is a pervasive problem in the remote sensing community. Accurate and efficient detection of cloud and cloud shadow is an essential step to harness the value of remotely sensed data for almost all downstream analysis. DeepMask, a new algorithm for cloud and cloud shadow detection in optical satellite remote sensing imagery, is proposed in this study. DeepMask utilizes ResNet, a deep convolutional neural network, for pixel-level cloud mask generation. The algorithm is trained and evaluated on the Landsat 8 Cloud Cover Assessment Validation Dataset distributed across 8 different land types. Compared with CFMask, the most widely used cloud detection algorithm, land-type-specific DeepMask models achieve higher accuracy across all land types. The average accuracy is 93.56\%, compared with 85.36\% from CFMask. DeepMask also achieves 91.02\% accuracy on all-land-type dataset. Compared with other CNN-based cloud mask algorithms, DeepMask benefits from the parsimonious architecture and the residual connection of ResNet. It is compatible with input of any size and shape. DeepMask still maintains high performance when using only red, green, blue, and NIR bands, indicating its potential to be applied to other satellite platforms that only have limited optical bands.
\end{abstract}

\section{Introduction}
\label{sec:introduction}

Detecting and masking out cloud and cloud shadow occlusion from satellite images is a classic and pervasive problem for the remote sensing community. Studies have found that mean annual global cloud coverage is estimated to be between 58\% \cite{rossow1999advances} and 66\% \cite{zhang2004calculation}. Cloud and cloud shadow contamination degrades the quality of remotely sensed data \cite{ju2008availability, li2017global} and increases the difficulty of further data processing (e.g. atmospheric correction) and all sorts of downstream remote sensing analysis, such as land type classification, target detection, and characterization of dynamics (e.g. forest, cropland, disaster). The problem is further complicated by the dynamic and diverse nature of clouds and land surfaces \cite{irish2000landsat}. Manual labeling of cloud and cloud shadow pixels are expensive in terms of time and human resources. Therefore, it is essential to develop high-quality, fully automated cloud mask algorithms.

Many cloud mask algorithms have been proposed to tackle this problem. These cloud mask algorithms can be divided into two major categories: multi-temporal algorithms and single-date algorithms. Multi-temporal algorithms utilizes temporal and statistical information to detect cloud pixels \cite{frantz2015enhancing, goodwin2013cloud, hagolle2010multi, zhu2018automatic, huang2010automated, hughes2014automated, jin2013automated, kennedy2010detecting, sheng2016representative, yang2003approach, zhu2018automatic, zhu2014automated}. Contaminated pixels are identified as statistical deviations from previous, temporally adjacent satellite acquisitions. The multi-temporal algorithms are usually computationally expensive (need to process large volume of time series data) and challenging to use (due to the requirement of a clear-sky reference image). 

Single-date algorithms can be further categorized into threshold-based approaches, machine learning approaches, and deep learning approaches. Threshold-based approaches use a set of predefined physical rules on spectral bands to detect clouds \cite{choi2004cloud, frantz2018improvement, helmer2009biomass, irish2000landsat, irish2006characterization, jin2013automated, oishi2018new, qiu2017improving, reeves2006fuels, roy2010web, scaramuzza2011development, vermote2007ledaps, zhu2012object, zhu2015improvement, qiu2019fmask}. Some approaches further derive cloud shadows from the detected clouds using the geometric relationship between the sun, the clouds, and the ground \cite{huang2010automated, qiu2017improving, vermote2007ledaps, zhu2015improvement, zhu2012object}. For all the threshold-based approaches, the exact values of the threshold tests are determined based on experiments and need to be manually tuned for new regions and equipment. Machine learning algorithms include fully-connected neural networks \cite{hughes2014automated, scaramuzza2011development, zi2018cloud}, fuzzy models \cite{melesse2002comparison, shao2017fuzzy}, and support vector machines \cite{zhou2016optional}. More recently, with the renaissance of deep learning and computer vision, several CNN-based cloud masks are proposed. A few studies process the input images using superpixel methods and then run a deep CNN on top of the superpixel representation to detect clouds \cite{xie2017multilevel, zi2018cloud}. However, these approaches rely on the superpixel pre-processing and the post-processing clean-up to generate decent results. A few others utilize the semantic segmentation networks for cloud mask generation \cite{chai2019cloud, isikdogan2017surface, liu2019clouds, wieland2019multi, zhaoxiang2018small}. To combine optical and temporal information, another study uses both CNN and RNN for cloud mask prediction \cite{tuia2018deep}. 

With various cloud mask algorithms available, USGS conducted a comprehensive analysis \cite{foga2017cloud} on those existing algorithms. The researchers collected and designed the Landsat 8 Cloud Cover Assessment (CCA) dataset to verify the effectiveness of those algorithms. According to their experiments, CFMask \cite{foga2017cloud}, a production C implementation by USGS EROS based on the Function of Mask (FMask)\cite{zhu2012object}, has the best overall performance. CFMask is a multi-stage algorithm that uses a pre-defined set of threshold tests to label pixels. It further refines cloud mask using statistics. It predicts cloud shadow using cloud height and satellite sensor projection. The CFMask algorithm is currently the most widely used cloud mask algorithm, and it is the default cloud mask algorithm used by Landsat missions. Even though the CFMask algorithm already achieves high-quality cloud mask, it is known to have several weaknesses: CFMask tends to confuse clouds with bright targets such as building tops, beaches, snow/ice, sand dunes, and salt lakes \cite{foga2017cloud, zhu2012object}. Optically thin clouds have a high probability of being omitted by the CFMask algorithm.

To improve the cloud and cloud shadow detection performance, we reflect on how human experts identify cloud and cloud shadow. Human experts analyze not only the spectral band values of each individual pixel, but also the shape and texture of the local region. This level of context and geospatial information is what CFMask and the other threshold-based approaches lack. Therefore, we share the same motivation of using Convolution Neural Networks (CNN) for cloud detection as some of the previous works did. We pick ResNet \cite{he2016deep}, one of the most widely-used CNN architectures, as the backbone and develop a new cloud mask algorithm, called “DeepMask”. Different from the previous CNN-based approaches, our model is single-stage and does not require any pre-processing or post-processing steps. Compared with the semantic segmentation and CNN-RNN networks, our algorithm aims for network parsimoniousness and efficiency while maintaining the same level of precision. Unlike any of the above-mentioned CNN-based approaches, which require the input images to have the same size, our algorithm accepts input of arbitrary size and shape. We evaluate our algorithm on the Landsat 8 CCA dataset, because of its comprehensive human label for cloud and cloud shadow mask and its extensive coverage of different land cover types.

When training and testing our algorithm, we develop a land-type-specific model (trained and tested on images from a specific land cover type) and an all-land-type model (trained and tested on images from all the land cover types). The latter one is aimed to test the generality of the DeepMask’s ability to detect cloud and shadow for images at different land cover types. Since we are using Landsat 8 in our case, we use the following spectral bands: band 1 (ultra-blue), band 2 (blue), band 3 (green), band 4 (red), band 5 (NIR), band 6 (SWIR 1), and band 7 (SWIR 2). We compare our performance against CFMask, following the convention of most cloud mask papers. To understand the contribution of different spectral bands to the cloud mask performance, we also conduct ablation experiments that use a subset of the spectral bands as input. The ablation experiments also test DeepMask’s potential to apply to satellites with fewer spectral bands. The current paper is organized as follows. In Section~\ref{sec:method}, we first introduce the data set and data preparation, and then explain the DeepMask algorithm, followed by experiment design and evaluation metrics. We present quantitative and qualitative results in Section~\ref{sec:results} and provide a detailed discussion of the results and future directions in Section~\ref{sec:discussion}. The conclusion is given in Section~\ref{sec:conclusion}.

\section{Materials and Methods}
\label{sec:method}
\subsection{Landsat 8 Cloud Cover Assessment Validation Dataset}
The Landsat 8 Cloud Cover Assessment (CCA) Validation dataset was created by Foga et al. from USGS for cloud detection algorithm validation. It contains 96 Landsat 8 Operational Land Imager (OLI) Thermal Infrared Sensor (TIRS) terrain-corrected (Level-1T) scenes. Manually generated cloud masks are used to validate cloud detection algorithms. Scenes were selected by 8 biomes defined by International Geosphere Biosphere Programme (IGBP) \cite{loveland2000development}: barren, forest, shrubland, cropland, snow/ice, wetlands, water, and urban. Compared with the selection of scenes by altitude, selection by biomes reduces the heterogeneity of land cover types within a given latitude zone. Twelve scenes were randomly selected from the path/row list of each biome to reduce clustering and maintain spatial as well as land-cover diversity.

\subsection{Data Preparation}
\label{preparation}
Human labeled cloud mask from the Landsat 8 CCA Level 1 dataset is used as our ground truth label (label band). We identify the product ID of each scene image and download the corresponding Landsat 8 Level 2 Surface Reflectance spectral bands. Unlike CFMask, DeepMask uses the atmospherically corrected surface reflectance bands (i.e. the fraction of radiation reflected by the earth’s surface within each spectral band) as input. We do not use thermal bands in our algorithm, which makes the algorithm compatible to other satellites without thermal sensors on-board. The spectral bands are at 30-meter resolution, and the size of each individual scene is approximately 170 km north-south and 183 km east-west. The CFMask performance is derived from the Quality Assessment (QA) band as our target of comparison.

To make the level 1 label band compatible with other level 2 spectral bands and QA band, the following procedures are applied: We first annotate the ground-truth label band with geolocation metadata from other level 1 bands from the CCA dataset. Then, we reproject the ground-truth label band to have the same extent $(x_{min}  ,y_{min},x_{max},y_{max})$ and origin as the corresponding level 2 bands. The valid pixel regions of label band and all other level 2 bands are slightly different. Valid pixel region is calculated as the intersection of the valid regions of level 1 and level 2 bands. This valid region is used to clip all spectral bands, the QA band, as well as the human label band, to maintain consistency.

\begin{figure}[t]
    \centering
    \includegraphics[width=0.6\columnwidth]{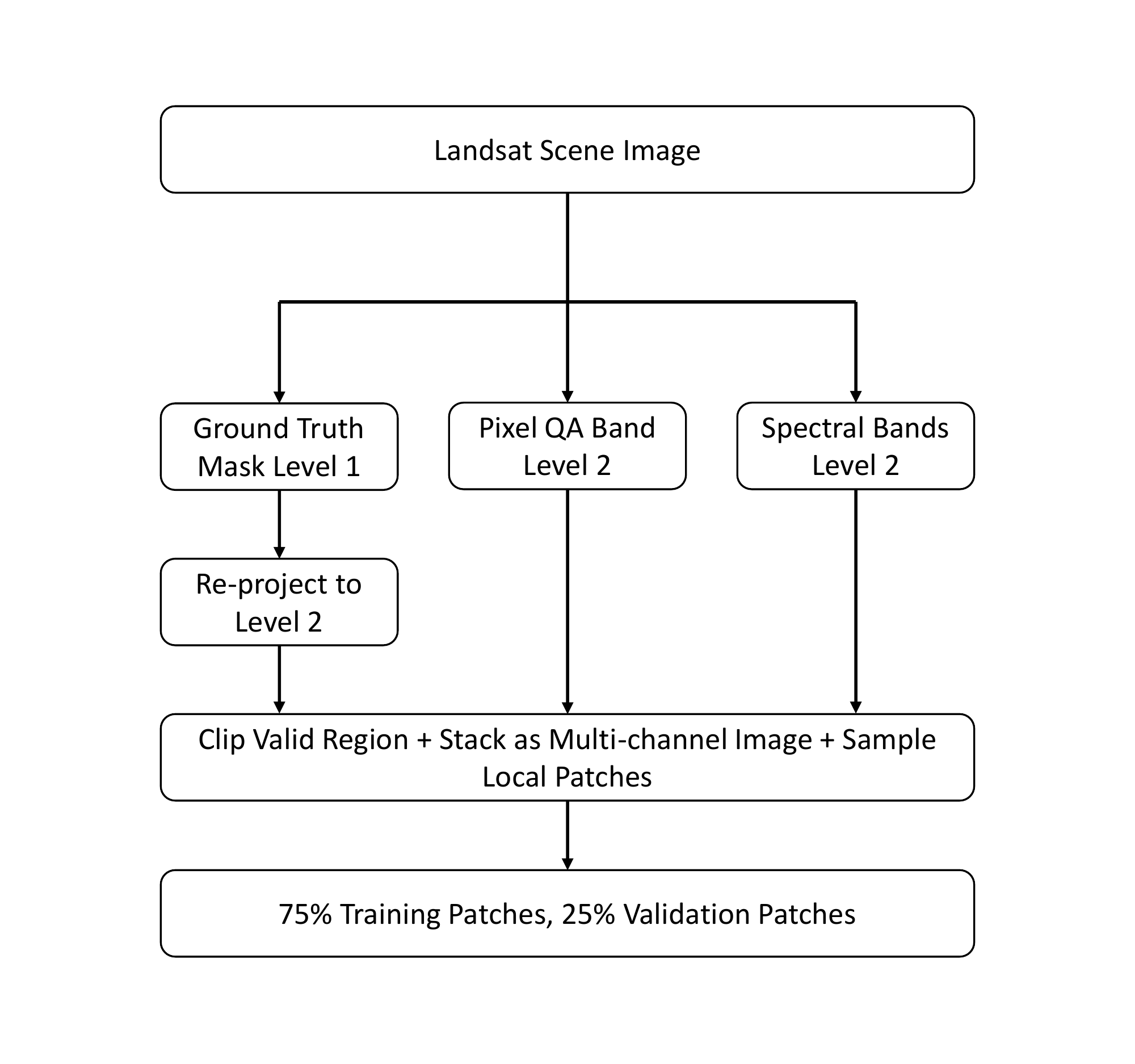}
    \caption{Data preparation flowchart}
    \label{fig:data_prep}
\end{figure}

We cast the cloud mask prediction as a binary classification problem. All pixels are divided into 2 classes: “clear” (ground) pixels and “cloud/shadow” pixels. We choose not to separate cloud and shadow pixels due to the high imbalance between cloud and shadow classes. The entire data preprocessing pipeline is summarized in Figure~\ref{fig:data_prep}.

\subsection{Convolutional Neural Network and ResNet}
\label{cnn}
Convolutional neural networks (CNNs) have led to advances in many computer vision applications, such as image classification \cite{he2016deep}, object detection \cite{redmon2016you}, and semantic segmentation \cite{ronneberger2015u}. CNNs are powerful at capturing local features and it has significantly fewer parameters to learn than a fully-connected neural network. By stacking multiple convolutional layers together, deep CNNs can integrate low-, mid- and high-level features in a hierarchical way. In this paper, we explore whether deep CNNs have the potential to be used for cloud and cloud shadow detection.

We pick ResNet \cite{he2016deep} as our deep CNN backbone. ResNet was first introduced in the 2015 ILSVRC \cite{russakovsky2015imagenet} competition and has become one of the most widely-used deep CNNs in recent years. ResNet features skip connections (module C of Figure~\ref{fig:model}) and the residual learning mechanism. For a target function $H(x)$, the network tries to learn a residual mapping $F(x)=H(x)-x$ instead of a direct mapping $F(x)=H(x)$. CNN architectures equipped with the shortcut connection have the same computational complexity as a plain CNN. ResNets can have much larger depth without the burden of the vanishing or exploding gradient problem.

Since the input local patch to our ResNet module is relatively small (15x15 pixels), we adopt the CIFAR-10 version of ResNet. The CIFAR-10 version of ResNet uses regular residual blocks instead of the bottleneck residual blocks. The first layer is a 3x3 convolution. The outputs are then passed to 3 residual blocks, each with \{16, 32, 64\} filters, respectively. The residual blocks are followed by an average pooling layer, a fully connected layer, and a softmax layer to give the corresponding probability for each class. The down-sampling of the output feature map is performed by the first convolution layer of each residual block. These convolution layers have a stride of 2 to reduce the output feature map size by half. All other convolutions have a stride of 1, which preserves the feature map size. This architecture results in a total of $6n+2$ weighted layers, where $n$ controls the depth of the ResNet. We experiment with ResNet of depth 20, 32, 44, and 56, which correspond to $n=\{3,5,7,9\}$ accordingly. The architectural details of ResNet 20 and ResNet 56 are provided in Table~\ref{tab:resnet}.

\begin{table}[h]
\setlength{\tabcolsep}{3pt}
\centering
\caption{ResNet 20 and 56 architecture. Residual blocks are shown in brackets, with the number of residual blocks shown outside the brackets. Within each bracket, filter size and number of filters are displayed.}
\label{tab:resnet}
\begin{tabular}{|c|c|c|c|}
\hline
\textbf{Layer Name} & \textbf{Feature Map Size} & \textbf{ResNet20} & \textbf{ResNet56} \\
\hline
\hline
conv1 & [15x15]x16 & 3x3, 16, stride 1 & 3x3, 16, stride 1 \\
\hline
resi block1 & [15x15]x16 & [conv3x3, 16]x6 & [conv3x3, 16]x18 \\
\hline
resi block2 & [8x8]x32 & [conv3x3, 32]x6 & [conv3x3, 32]x18 \\
\hline
resi block3 & [4x4]x64 & [conv3x3, 64]x6 & [conv3x3, 64]x18 \\
\hline
\multicolumn{4}{|c|}{average poll, 2-d fc, softmax} \\
\hline
\end{tabular}
\end{table}

\subsection{The DeepMask Algorithm}
\begin{figure*}[ht]
\centering
\includegraphics[width=0.98\textwidth]{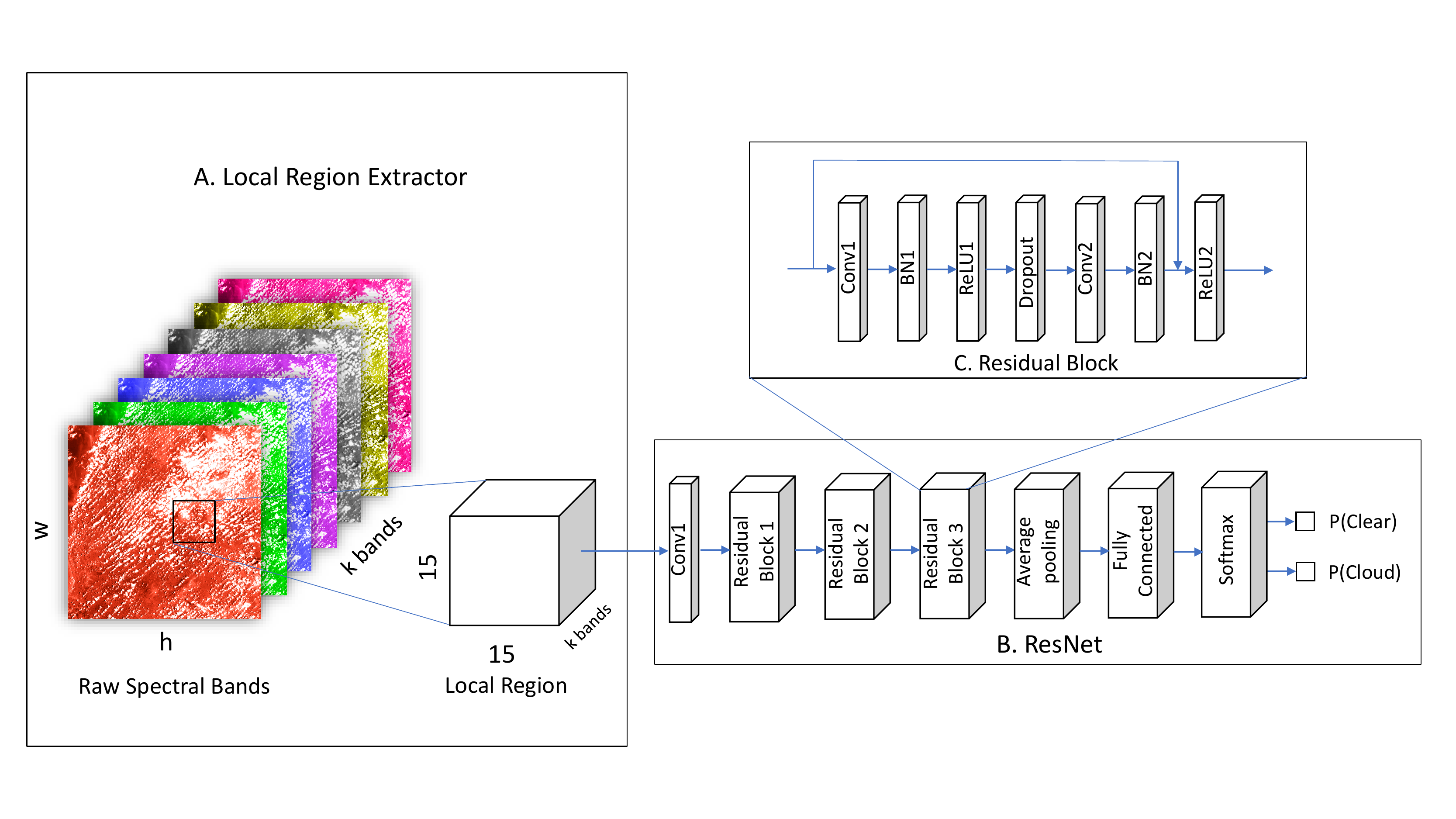}
\caption{DeepMask algorithm is a unified pipeline, consisting of the local region extractor (module A) and the ResNet backbone (module B). A zoomed-in view (module C) of a typical residual block is also given.}
\label{fig:model}
\end{figure*}

Our cloud and cloud shadow detection algorithm, called DeepMask, is a unified training and testing pipeline that uses ResNet as the backbone. Figure~\ref{fig:model} gives an overview of the DeepMask algorithm. DeepMask consists of two modules. The first module is the local region extractor (module A), and the second module is the ResNet central pixel classifier (module B). The inputs to the DeepMask algorithm are raw spectral bands of a satellite scene image of arbitrary size and shape. Here we use Landsat 8 images for illustration. The local region extractor takes the raw spectral bands (shown as a stack of image bands in different colors) and extracts 15x15 pixel local regions. We extract local regions because feeding the entire scene image into a deep CNN will be too computationally expensive for high-resolution satellite images. The various sizes of satellite images is another reason, since ResNet requires fixed input sizes. For a 30-meter resolution Landsat 8 scene image, the local region gives a 450-meter by 450-meter raw image. The ResNet module then takes the 15x15 pixel local region and predicts the class label for the central pixel of the local patch. A zoomed-in view of a typical residual block is given in module C. The final output of the ResNet module is the binary probability distribution of the central pixel class membership (Clear, Cloud/Shadow). This probability can be interpreted as the cloud confidence level of the algorithm. However, the cloud confidence level should not be misinterpreted as cloud thickness, since cloud confidence depends not only on the cloud's thickness but also on other factors (e.g. spatial heterogeneity, texture, etc.). Therefore, this cloud confidence level can be used to balance the precision-recall trade-off but should not be used to filter thick clouds and leave thin clouds. 

During training, the local region extractor subsamples local patches from the training dataset. This step reduces the computational cost of training the ResNet classifier compared with using all the local regions for each image in the training set. During testing, since we would like to predict the class label for all valid pixels in a test scene image of arbitrary size and shape, the local region extractor acts as a sliding window. It iterates through all the valid local regions in the scene and passes the extracted local regions to the trained ResNet to predict the corresponding label for the central pixel. In the entire DeepMask pipeline, only the ResNet module needs to be trained. The loss function for training the ResNet model is the binary class cross entropy loss:
\begin{equation}
    -\frac{1}{N} \sum_{i=1}^N [y_i \log{p(y_i)} + (1-y_i)\log{(1-p(y_i))}]
\end{equation}
where $N$ is the mini-batch size, $y_i$ is the class label of the i-th data sample (0 for a clear central pixel and 1 for a cloud/shadow central pixel) and $p(y_i )$ is the predicted softmax probability of the central pixel being a cloud/shadow pixel.

\subsection{Experiment Design}

\subsubsection{Land-type-specific Experiments}
In land-type-specific experiments, we train a model for each land cover type and evaluate its performance on images of that specific land type. Concretely, for each land type, we use “leave-one-out” training and testing scheme. For example, “barren” has 12 scene images. One scene image is left out as the test image, and all other 11 images are used for training and validation. We iterate for 12 times by selecting each one out of 12 images as the test image. Thus, for each land cover type, we have trained 12 models, which will be tested on 12 different scene images. The test performance for that land type “barren” will be the average of those 12 test results. The advantage of the leave-one-out approach, as compared to directly reserving a fraction of pixels from all scenes for out-of-sample validation, is that it tests the model’s generalization performance on scenes that it has never seen during training. Since the within-scene heterogeneity is somewhat smaller than the cross-scene difference, leave-one-out is stricter than reserving a fraction of all pixels from all scenes.

As shown in Figure~\ref{fig:model}, the residual network takes a 15-by-15 local region as input and predicts the class label of the central pixel. A 15-by-15 local region is considered “valid” if it does not contain any invalid pixel. For training and validation efficiency, we subsample 10,000 valid local regions from each training image. To ensure no overlap between training local regions and validation local regions, we divide each scene image into a 2x2 grid of sub-images. We sampled 2,500 local regions from each sub-image. Local patches from one of those sub-images are used as validation data and local patches from the other 3 sub-images are used as training data. The sub-image that is left out as the validation set is assigned randomly. Following this scheme, 75\% of the local regions are used as training data and 25\% are used as validation data, as shown in the last step of Figure~\ref{fig:data_prep}. On average, each land type has around 90,000 training patches and 30,000 validation patches.

For training ResNet, we have experimented with ResNet 20, 32, 44, and 56, and ResNet 20 gives the best results. We use SGD with nesterov momentum (Nesterov, 1983) and a mini-batch size of 256. The learning rate starts from 0.1 and is reduced by a factor of 10 whenever the validation loss plateaus. The ResNet module is typically trained for 80-120 epochs. We use an L2 weight decay of 0.0005 and a Dropout of 0.5.

During testing, we do not subsample local regions. All valid 15-by-15 local regions of each test scene image are used. The trained model will slide through each valid local region and predict the class label of the central pixel. A complete cloud mask for each test image is produced. As a reference for the inference speed, each test image takes about 7-8 mins on a Desktop computer with Intel i7 CPU and one NVIDIA 1080 Ti GPU.

\subsubsection{All-land-type Experiments}
\label{sec:alllandtype}
We also apply DeepMask to all the land cover types, aiming to test how generic DeepMask can be when trained using images from all the land cover types. Different from the land-type-specific experiments, all-land-type models are trained using a sample of images from all the land cover types, and then tested on a set of reserved test images.  During the training stage, for each land type, we randomly reserve one image as the test image and use all other images for training and validation. This gives us 8 images for testing and 78 images for training and validation. We use 20\% of those images from the training-validation pool as validation set and 80\% as training set. We mix the training and validation set of different land types together and randomly subsample 10,000 valid location regions from each image.  All the remaining training components are the same as the land-type-specific experiments.

During the testing stage, we evaluate the performance on those 8 leave-one-out test images. To reduce experimental randomness and provide a better evaluation of our model, we repeat the experiment five times. Each time we resample all the training data and select the leave-one-out test image randomly. The overall all land type performance is the average performance across those five experiments.

\subsubsection{Ablation Experiments}
To analyze the contribution of individual spectral band to cloud and shadow detection accuracy, we also perform ablation experiments on the spectral bands. These experiments also give insights to the transferability of our algorithm to other satellites with fewer spectral bands. We first analyze the effect of dropping a single spectral band to the overall cloud detection performance. We iteratively drop one of the 7 spectral bands (ultra-blue, blue, green, red, NIR, SWIR 1 and SWIR 2) from the input local patch and train the detection network. That is, the input local region now has 6 channels instead of 7. The test images also have the corresponding band dropped. All the remaining training, validation, and testing mechanisms are the same as all land type experiments explained in Section~\ref{sec:alllandtype}. To account for satellites like PlanetScope from Planet Lab, we also perform an ablation experiment with only 4 basic spectral bands: red, green, blue, and NIR. Similar to the all-land-type experiments, for each of the “drop-single-band” and “keep-4-bands” experiments, we repeat the experiment five times and take the average performance.

\subsection{Evaluation Metrics}
Following the tradition in the cloud mask and machine learning literatures, we select overall accuracy, precision, recall, F1 score, Area Under the Receiver Operating Characteristic Curve (AUROC), and Average Precision (AP) as our evaluation metrics. We adjust the metrics definitions slightly for them to be compatible with our binary classification setup. Accuracy, AUROC, and AP are calculated for both clear and cloud/shadow class, while precision, recall, and F1 score are calculated for each class separately. The detailed definitions of evaluation metrics are given in the \textit{appendix section}.

\section{Results}
\label{sec:results}

\subsection{Quantitative Results}

\subsubsection{Land-type-specific Experiments}
We first present the performance of our land-type-specific model. The average performance across all land types is given in Table~\ref{tab:landtype_spec_avg} and the per-land-type performance is given in Table~\ref{tab:pertype_deepmask}. Precision, recall, and F1 score are calculated for each class separately. All other metrics are calculated for both classes.
According to Table~\ref{tab:landtype_spec_avg}, the overall accuracy of the DeepMask algorithm is 93.56\%, which is 8.2\% higher than the 85.36\% overall accuracy of CFMask. Both clear and cloud/shadow precisions of DeepMask are higher than that of CFMask, indicating DeepMask is reliable in the prediction it makes. The cloud recall for DeepMask is 1.35\% lower, indicating that under 50\% confidence threshold, DeepMask is slightly worse in terms of finding all the true positive for cloud. However, this sensitivity vs. specificity can be adjusted using a different threshold. Users can adjust how conservative the cloud mask is by varying the cloud confidence threshold. To further explore this trade-off, we also provide AUROC and AP for DeepMask. We do not calculate these two metrics for CFMask, since a 0 to 1 prediction confidence is not available from the QA band.

\begin{table*}[h]
\setlength{\tabcolsep}{2.5pt}
\centering
\caption{Land-type-specific model test performance averaged across all land types}
\label{tab:landtype_spec_avg}
\begin{adjustbox}{width=\textwidth}
\begin{tabular}{|c|c|c|c|c|c|c|c|c|c|c|}
\hline
& \textbf{Accuracy} & \textbf{Precision Clear} & \textbf{Precision Cloud} & \textbf{Recall Clear} & \textbf{Recall Cloud} & \textbf{F1 Clear} & \textbf{F1 Cloud} & \textbf{AUROC} & \textbf{AP} \\
\hline
\hline
\textbf{DeepMask} & 93.56\% & 93.85\% & 94.76\% & 94.10\% & 92.80\% & 93.66\% & 93.42\% & 93.78\% & 87.72\% \\
\hline
\textbf{CFMask} & 85.36\% & 91.02\% & 88.09\% & 76.89\% & 94.15\% & 81.85\% & 89.49\% & - & - \\
\hline
\end{tabular}
\end{adjustbox}
\end{table*}

We also present the performance of DeepMask as well as CFMask for each land type, as shown in Table~\ref{tab:pertype_deepmask} and Table~\ref{tab:pertype_cfmask} respectively. CFMask is known to have poor performance on snow/ice land types, due to the bright surface and similar spectral signals between snow/ice and cloud. According to the results, DeepMask performs much better than CFMask in snow/ice land cover types, with 24.98\% higher accuracy. For barren, shrubland, urban, water, and wetland, DeepMask achieves 6-8\% higher accuracy than CFMask. DeepMask has similar performance with CFMask on crops and forest, but slightly better. Overall, DeepMask shows better cloud/shadow detection capability in all 8 different land cover types.

\begin{table*}[h]
\setlength{\tabcolsep}{2.5pt}
\centering
\caption{Land-type-specific DeepMask test performance by land type}
\label{tab:pertype_deepmask}
\begin{adjustbox}{width=\textwidth}
\begin{tabular}{|c|c|c|c|c|c|c|c|c|c|c|}
\hline
& \textbf{Accuracy} & \textbf{Precision Clear} & \textbf{Precision Cloud} & \textbf{Recall Clear} & \textbf{Recall Cloud} & \textbf{F1 Clear} & \textbf{F1 Cloud} & \textbf{AUROC} & \textbf{AP} \\
\hline
\hline
\textbf{Barren} & 94.70\% & 95.66\% & 94.83\% &	93.40\% & 95.90\% & 94.13\% & 95.18\% & 96.88\% & 96.87\% \\
\hline
\textbf{Crops} & 94.13\% & 94.43\% & 95.47\% & 94.49\% & 93.76\% & 94.17\% & 94.04\% & 94.08\% & 82.38\% \\
\hline
\textbf{Forest} & 92.28\% & 89.41\% & 95.54\% & 92.12\% & 92.37\% & 90.00\% & 93.32\% & 92.46\% & 94.49\% \\
\hline
\textbf{Shrubland} & 93.02\% & 93.67\% & 95.15\% & 94.25\% & 91.77\% & 93.52\% & 93.19\% & 93.56\% & 85.04\% \\
\hline
\textbf{Snow} & 86.42\% & 89.02\% & 88.49\% & 90.21\% & 81.00\% & 89.28\% & 84.02\% & 89.66\% & 76.86\% \\
\hline
\textbf{Urban} & 96.05\% & 96.43\% & 96.16\% & 96.30\% & 95.77\% & 96.28\% & 95.54\% & 89.20\% & 81.91\% \\
\hline
\textbf{Water} & 95.77\% & 96.25\% & 95.67\% & 95.71\% & 95.84\% & 95.89\% & 95.66\% & 96.31\% & 90.47\% \\
\hline
\textbf{Wetland} & 96.15\% & 95.92\% & 96.81\% & 96.30\% & 96.02\% & 96.03\% & 96.40\% & 98.08\% & 93.75\% \\
\hline
\end{tabular}
\end{adjustbox}
\end{table*}

\begin{table*}[h]
\setlength{\tabcolsep}{2.5pt}
\centering
\caption{Land-type-specific CFMask test performance by land type}
\label{tab:pertype_cfmask}
\begin{adjustbox}{width=\textwidth}
\begin{tabular}{|c|c|c|c|c|c|c|c|}
\hline
& \textbf{Accuracy} & \textbf{Precision Clear} & \textbf{Precision Cloud} & \textbf{Recall Clear} & \textbf{Recall Cloud} & \textbf{F1 Clear} & \textbf{F1 Cloud} \\
\hline
\hline
\textbf{Barren} & 88.79\% & 91.64\% & 87.94\% & 81.18\% & 95.85\% & 84.40\% & 90.82\% \\
\hline
\textbf{Crops} & 91.28\% & 96.82\% & 92.07\% & 85.72\% & 96.97\% & 89.87\% & 93.58\% \\
\hline
\textbf{Forest} & 91.43\% & 93.80\% & 90.68\% & 80.94\% & 97.63\% & 85.61\% & 93.74\% \\
\hline
\textbf{Shrubland} & 84.83\% & 95.20\% & 86.71\% & 83.53\% & 86.16\% & 87.52\% & 81.95\% \\
\hline
\textbf{Snow} & 61.44\% & 56.26\% & 85.98\% & 44.89\% & 85.13\% & 47.18\% & 83.36\% \\
\hline
\textbf{Urban} & 89.06\% & 99.33\% & 86.60\% & 79.99\% & 99.41\% & 86.78\% & 91.55\% \\
\hline
\textbf{Water} & 87.61\% & 96.70\% & 87.41\% & 83.11\% & 92.55\% & 88.57\% & 88.31\% \\
\hline
\textbf{Wetland} & 88.44\% & 98.43\% & 87.31\% & 75.72\% & 99.55\% & 84.85\% & 92.59\% \\
\hline
\end{tabular}
\end{adjustbox}
\end{table*}

\subsubsection{All-land-type and Ablation Experiments}
To test the generalizability of the DeepMask algorithm, we also conduct all-land-type experiments (Table~\ref{tab:all_type}), in which we train and test on all land type data. Compared with a land-type-specific model, we observe a slight decrease in accuracy as well as F1-score. This indicates the trade-off between a general, less accurate model and a problem-specific, more accurate model. 

In the drop-single-band experiments, dropping red, blue, ultra-blue or swir1 results in a 3-4\% decrease in accuracy. This observation is consistent with the domain knowledge that blue band is sensitive to cloud but not sensitive to vegetation \cite{liu2016reference, govender2007review}. Dropping other bands results in only a slight decrease in performance.

We also trained a model with only 4 bands: red, green, blue, and NIR. This results in a model with only a 1\% decrease in performance. The results demonstrate the applicability of DeepMask to lower-cost CubeSat missions with limited spectral coverage, such as Planet Lab’s PlanetScope fleet. In all cases, DeepMask can achieve better accuracy than CFMask in terms of cloud and cloud shadow detection.

\begin{table*}[h]
\setlength{\tabcolsep}{2.5pt}
\centering
\caption{All land type model test performance and ablation experiments}
\label{tab:all_type}
\begin{adjustbox}{width=\textwidth}
\begin{tabular}{|c|c|c|c|c|c|c|c|c|c|c|}
\hline
& \textbf{Accuracy} & \textbf{Precision Clear} & \textbf{Precision Cloud} & \textbf{Recall Clear} & \textbf{Recall Cloud} & \textbf{F1 Clear} & \textbf{F1 Cloud} & \textbf{AUROC} & \textbf{AP} \\
\hline
\hline
\textbf{All bands} & 91.02\% & 92.62\% & 95.47\% & 89.88\% & 90.47\% & 88.81\% & 91.78\% & 94.68\% & 87.86\% \\
\hline
\textbf{Drop ultra-b} & 89.69\% & 90.84\% & 93.89\% & 87.68\% & 89.27\% & 87.09\% & 89.88\% & 93.20\% & 87.75\% \\
\hline
\textbf{Drop R} & 88.88\% & 90.99\% & 95.04\% & 88.65\% & 88.51\% & 87.49\% & 89.96\% & 91.15\% & 88.02\% \\
\hline
\textbf{Drop G} & 90.64\% & 91.37\% & 95.24\% & 90.85\% & 88.96\% & 89.26\% & 90.31\% & 92.42\% & 86.16\% \\
\hline
\textbf{Drop B} & 89.60\% & 91.71\% & 95.51\% & 88.52\% & 88.46\% & 87.07\% & 89.83\% & 93.88\% & 87.47\% \\
\hline
\textbf{Drop NIR} & 90.62\% & 92.84\% & 94.21\% & 89.30\% & 90.02\% & 89.53\% & 91.06\% & 93.00\% & 86.73\% \\
\hline
\textbf{Drop SWIR1} & 89.62\% & 92.34\% & 94.75\% & 88.91\% & 89.99\% & 88.37\% & 90.98\% & 92.85\% & 87.97\% \\
\hline
\textbf{Drop SWIR2} & 90.37\% & 93.00\% & 94.60\% & 89.31\% & 89.87\% & 89.24\% & 90.60\% & 93.81\% & 87.64\% \\
\hline
\textbf{R, G, B, NIR} & 90.01\% & 90.20\% & 93.72\% & 88.25\% & 90.37\% & 87.08\% & 91.04\% & 92.75\% & 87.01\% \\
\hline
\textbf{CFMask} & 85.80\% & 91.23\% & 88.18\% & 75.30\% & 95.20\% & 80.73\% & 90.40\% & - & - \\
\hline
\end{tabular}
\end{adjustbox}
\end{table*}

\subsection{Qualitative Results}
In addition to the numerical results, we provide visualizations for each of the 8 land-type-specific models to visually assess DeepMask’s performance, as shown in Figure 3 and Figure 4. For each land cover type, we select one typical scene image. The location, cloud ratio, and other relevant information of the test scenes are provided in Table~\ref{tab:scene_info}.

\begin{table*}[t]
\setlength{\tabcolsep}{2.5pt}
\centering
\caption{Information of scenes in visualization}
\label{tab:scene_info}
\begin{adjustbox}{width=\textwidth}
\begin{tabular}{|c|c|c|c|c|c|}
\hline
\textbf{Land Type} & \textbf{WRS Path} & \textbf{WRS Row} & \textbf{Location} & \textbf{Date} & \textbf{Cloud Ratio} \\
\hline
\hline
Snow/Ice & 100 & 108 & Ice shield, Antarctica & 1/22/2014 & 3.20\% \\
\hline
Water & 65 & 18 & South Alaska, US & 8/25/2013 & 51.32\% \\
\hline
Crops & 132 & 35 & Hainan, Qinghai, China & 8/31/2013 & 32.41\% \\
\hline
Forest & 133 & 18 & Lenskiy Ulus, Sakha Republic, Russia & 7/5/2013 & 97.63\% \\
\hline
Wetland & 146 & 16 & Krasnoyarsk Krai, Russia & 6/17/2014 & 0.00\% \\
\hline
Barren & 157 & 45 & Ash Sharqiyah South Governorate, Oman & 8/1/2014 & 61.91\% \\
\hline
Shrubland & 159 & 36 & Razavi Khorasan Province, Iran & 2/20/2014 & 51.82\% \\
\hline
Urban & 162 & 43 & Persian Gulf & 3/13/2014 & 20.29\% \\
\hline
\end{tabular}
\end{adjustbox}
\end{table*}

We display the RGB scene image, ground-truth label mask, CFMask cloud mask, and DeepMask cloud mask. The RGB image is generated by combining the red, green, and blue bands of the image. Ground-truth label mask, CFMask cloud mask and DeepMask cloud mask are displayed as discrete color maps. Gray represents clear (ground) pixels and white represents cloud/shadow pixels. 

\begin{figure*}[h]
\centering
\includegraphics[width=0.7\textwidth]{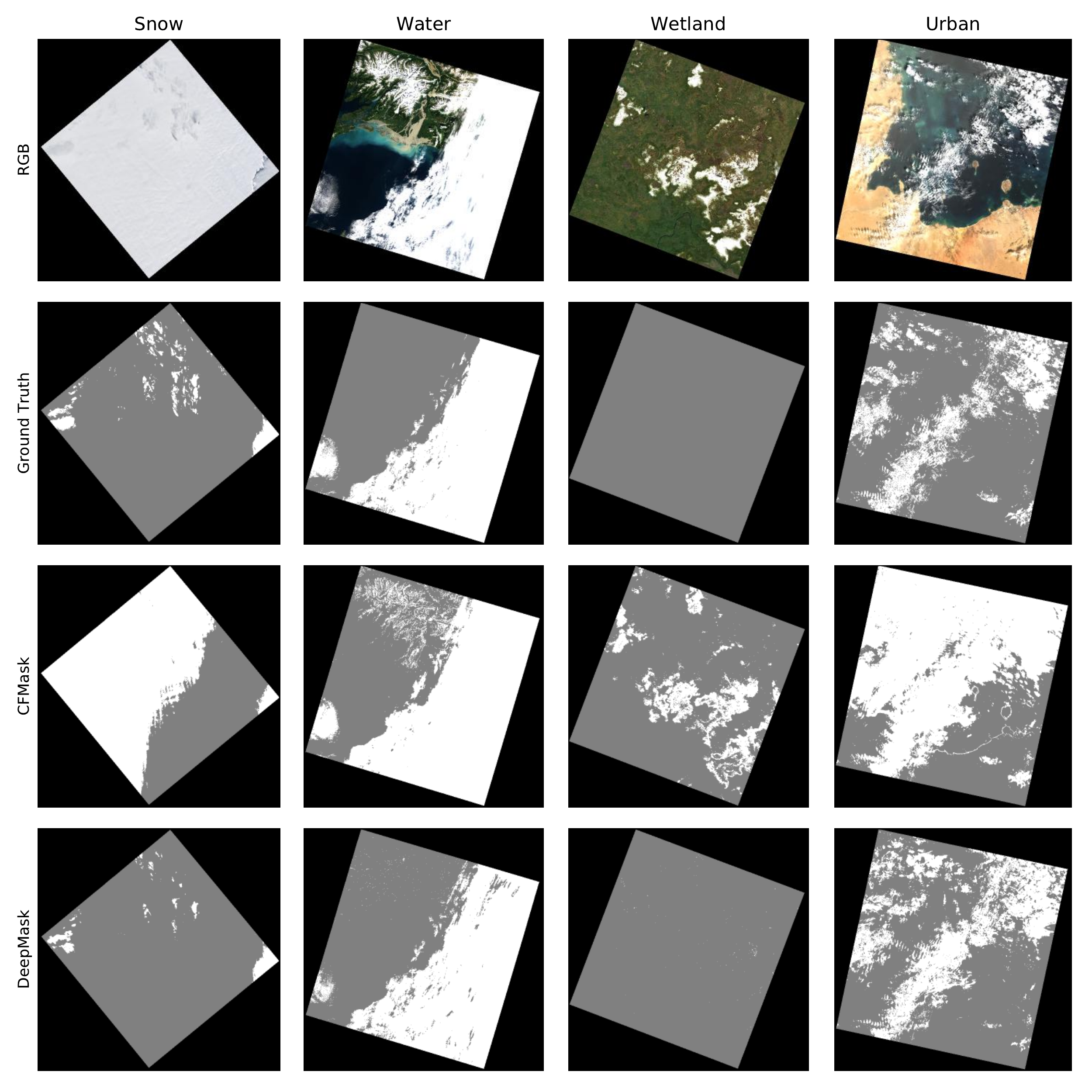}
\caption{Example visualization of raw RGB image (1st row), ground truth labels (2nd row), CFMask results (3rd row), and DeepMask results (last row), for four land types: snow (1st column), water (2nd column), wetland (3rd column) and urban (last column). For ground truth label, CFMask results and DeepMask results, gray color denotes clear pixels, and white color denotes cloud/shadow pixels.}
\label{fig:visual1}
\end{figure*}

The \textbf{1st} column of Figure~\ref{fig:visual1} corresponds to an ice shelf image from Antarctica. Most of the scene is covered with snow and ice, with only scattered cloud. DeepMask shows a strong ability to distinguish between snow/ice and cloud/shadow. CFMask, on the other hand, overestimates the cloud/shadow and is not able to separate cloud from snow. The \textbf{2nd column} corresponds to a mixture of ocean water, green land, snow and cloud located at South Alaska, US. The similarity in spectral patterns of snow, ice, water and cloud make it a challenging image. CFMask incorrectly classifies snow/ice on mountains as cloud, while DeepMask is able to identify snow/ice and locate cloud location precisely. The \textbf{3rd column} corresponds to a typical image of wetland in Krasnoyarsk Krai, Russia with snow on top of the mountains. The spectral signal and shape of the snow resemble those of the cloud. Base on the ground truth manual mask, this scene is cloud-free. However, CFMask mistakenly classifies all snow as cloud. DeepMask also makes mistakes in some scattered bright spots, but the performance is much better. The \textbf{last column} corresponds to a sub-urban image in the Persian Gulf with thick and thin clouds over bright urban surfaces. DeepMask is able to precisely identify cloud/shadow, although it overestimates slightly. CFMask, on the other hand, overestimates a large portion of water pixel as thin cloud. It also mistakenly classifies some bright island and peninsula contours as cloud, as shown in the lower-right area of the CFMask cloud mask.

\begin{figure*}[h]
\centering
\includegraphics[width=0.7\textwidth]{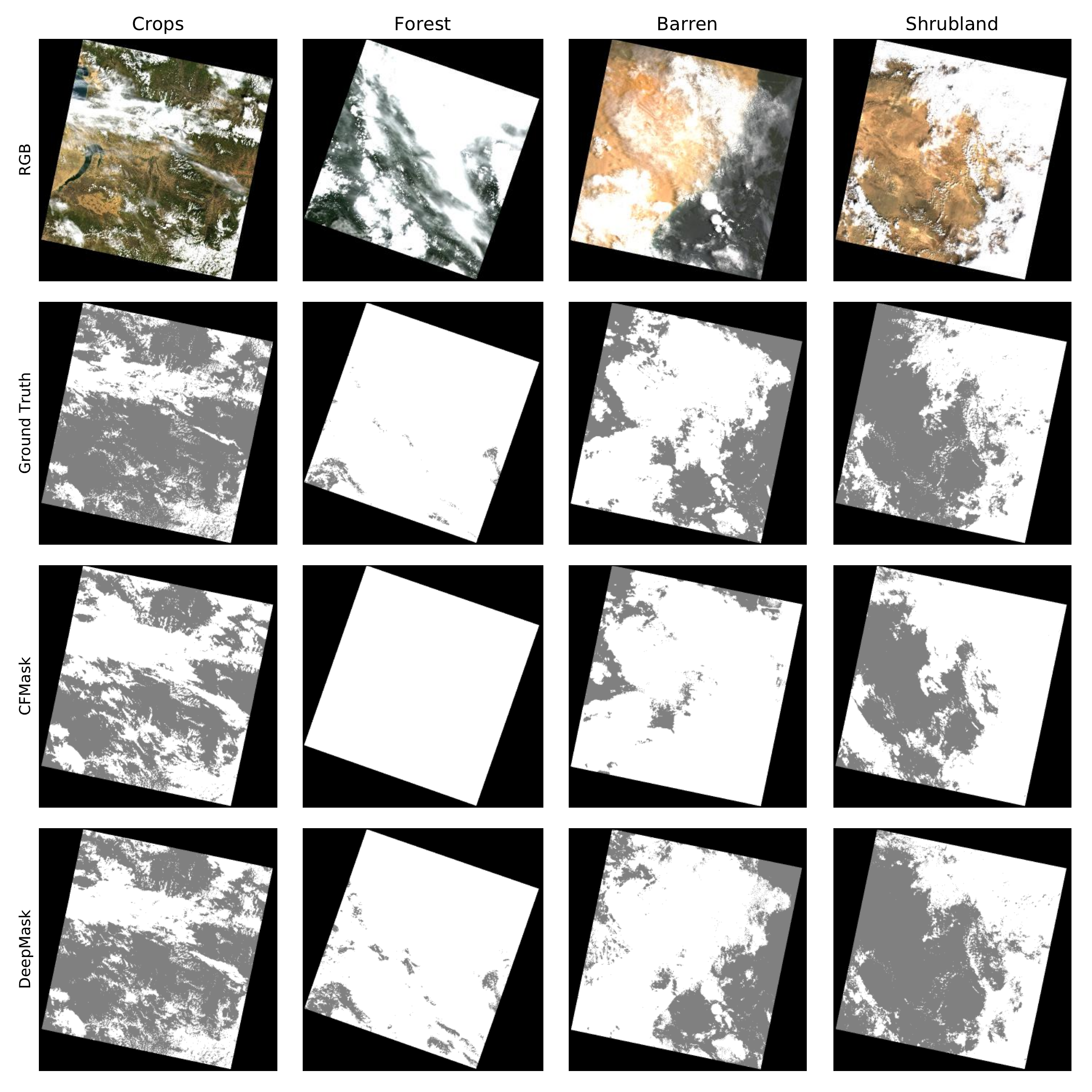}
\caption{Example visualization of raw image (RGB), ground truth labels, CFMask results, and DeepMask results, for four land types: crops, forest, barren and shrubland. For ground truth label, CFMask results and DeepMask results, gray color denotes clear pixels, and white color denotes cloud/shadow pixels.}
\label{fig:visual2}
\end{figure*}

The \textbf{1st column} of Figure~\ref{fig:visual2} corresponds to one of the typical crop land images. It is located in Qinghai, China. 32.41\% of the scene is covered with thick and thin cloud. Both CFMask and DeepMask are able to locate most cloud pixels, although both of them overestimate the cloud extent. DeepMask performs slightly better, since it has less overestimated cloud pixels. The \textbf{2nd column} corresponds to one of the cloudier images. It is a forest image located in Lenskiy Ulus, Sakha Republic, Russia. Although most of the image is covered with cloud, it does have some clear pixels, which are identified by DeepMask. CFMask simply predict all pixels as cloud and is thus less effective. The \textbf{3rd column} corresponds to a mixture of water and bright barren land in Ash Sharqiyah South Governorate, Oman. Water and water vapor pixels resemble the spectral signal of cloud and thus confuse CFMask. Although it overestimates slightly, DeepMask is able to identify the boundary between ocean and barren. The \textbf{last column} corresponds to a typical shrubland image from Razavi Khorasan Province, Iran. Although both DeepMask and CFMask make some minor misclassifications, both perform well in this scene image.

\section{Discussion}
\label{sec:discussion}

In this study, we present DeepMask as a highly accurate cloud and cloud shadow mask algorithm. Benchmarked on the CCA database with all spectral bands, the land-type-specific DeepMask model is able to outperform CFMask in all 8 different land cover types. The high performance of DeepMask comes from deep CNN’s ability to utilize the textual as well as the spatial and geometric information, in additional to individual band values. Deep CNN also integrates low-, medium-, and high-level visual features in a hierarchical way. Different from the other deep CNN networks, the residual connections in ResNet make it more efficient and powerful. DeepMask also benefits from the local region extraction mechanism, making prediction for each individual central pixel rather than applying the same threshold to all pixels, as in threshold-based approaches. CFMask is known to perform poorly over water, snow/ice land types. DeepMask, on the other hand, is able to well distinguish cloud/shadow from snow/ice. 

DeepMask is a flexible algorithm with sufficient generality. The all-land-type model of DeepMask results in a less than 1\% decrease in accuracy compared to the land-type-specific version. However, the all-land-type model of DeepMask still outperforms CFMask by a large margin, indicating its ability to generalize to different biomes and surface textures. DeepMask gives the user the freedom to use the algorithm in a local region or at a larger scale. We purposely did not include thermal bands during training, which makes our model applicable to satellites without a thermal band. According to the ablation experiments, DeepMask shows no strong sensitivity in its performance when dropping any specific spectral bands. DeepMask using only 4 bands still achieves similarly high performance, indicating its potential applications to satellites with fewer spectral bands.

One side effect of using a 15x15 pixel local region as input is that it will result in a 7-pixel border of the image that cannot be classified. However, all CNN-based algorithm will have this issue, which can be solved using reflection or replication padding. To apply DeepMask algorithm to Landsat 7, which is known to have the gap issue due to Scan Line Corrector (SLC) failure, morphological dilation or other gap-filling techniques \cite{zheng2012remote, potapov2011regional, roy2010web} can be used first before applying the DeepMask algorithm. On the other hand, the local region extraction and sliding window inference mechanism allow DeepMask to handle input of various sizes and shapes.

DeepMask has the potential to be improved through future efforts. The efforts may include collecting more manually labeled data and training a land-type-specific model in some land types that currently CFMask and DeepMask have similar performance (e.g. crop and forest). Incorporating temporal and thermal signals may further improve the cloud detection performance. Further optimizing computational cost would be another direction of improvement. To facilitate the community’s use and further development, we will share our code with examples for open-access uses at: \url{https://www.richardkxu.com/dl-remote-sensing}.

\section{Conclusion}
\label{sec:conclusion}
This paper presents DeepMask, an algorithm developed based on ResNet for cloud and cloud shadow mask generation for Landsat 8 imagery, but potentially widely applicable to different sorts of optical satellite imagery. Compared with the threshold-based methods, DeepMask utilizes both spectral signals as well as 2D visual cues, resulting in a large increase in detection performance. Compared with the other CNN-based cloud masks, DeepMask is simpler, more efficient, and more flexible with input of various sizes and shapes. DeepMask algorithm can be used in land-type-specific problems as well as all-land-type problems. It provides options to deal with the trade-off between precision and recall by changing the cloud confidence threshold. It is compatible with satellites without thermal bands and with fewer spectral bands. DeepMask is a new promising approach in the field of cloud and cloud shadow detection for optical satellite remote sensing.

\section*{Acknowledgment}
We acknowledge the support of DOE CABBI. Guan also acknowledge the support from NASA Terrestrial Ecology Program for the NASA Carbon Monitoring System program and NASA New Investigator Program. This research is part of the Blue Waters sustained-petascale computing project, which is supported by the National Science Foundation (awards OCI-0725070 and ACI-1238993) and the state of Illinois. Blue Waters is a joint effort of the University of Illinois at Urbana-Champaign and its National Center for Supercomputing Applications.

\bibliographystyle{abbrv}
\bibliography{IEEEabrv,IEEEexample, mybib}

\end{document}